\documentclass{article}

\usepackage{arxiv}

\usepackage[utf8]{inputenc} 
\usepackage[T1]{fontenc}    
\usepackage{hyperref}       
\usepackage{url}            
\usepackage{booktabs}       
\usepackage{amsfonts}       
\usepackage{nicefrac}       
\usepackage{microtype}      
\usepackage{lipsum}		
\usepackage{graphicx}
\usepackage{doi}
\usepackage{multirow}
\usepackage[dvipsnames]{xcolor}

\title{Surgical Action Triplet Detection by Mixed Supervised Learning of Instrument-Tissue Interactions}

\author{ 
        {
        \hspace{12mm}\href{https://orcid.org/0000-0002-6021-6132}{\includegraphics[scale=0.06]{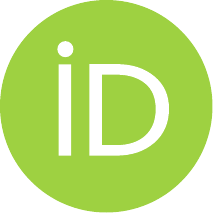}} Saurav Sharma} \\
	{\hspace{14mm}ICube Laboratory}\\
	{\hspace{10mm}University of Strasbourg, France}\\
	{\hspace{14mm}\texttt{ssharma@unistra.fr}} \\
	\And
	{
        \hspace{23mm}\href{https://orcid.org/0000-0003-4777-0857}{\includegraphics[scale=0.06]{orcid.pdf}} Chinedu Innocent Nwoye} \\
	{\hspace{22mm}ICube Laboratory}\\
	{\hspace{22mm}University of Strasbourg, France}\\
	{\hspace{22mm}\texttt{nwoye@unistra.fr}}\\
	\And
        \And
 	{
         \hspace{1mm}\href{https://orcid.org/0000-0002-7559-3328}{\includegraphics[scale=0.06]{orcid.pdf}} Didier Mutter} \\
        IHU Strasbourg, France\\
        University Hospital of Strasbourg, France\\
	\texttt{didier.mutter@ihu-strasbourg.eu} \\
        \And
	{
        \hspace{1mm}\href{https://orcid.org/0000-0002-5010-4137}{\includegraphics[scale=0.06]{orcid.pdf}} Nicolas Padoy} \\
        IHU Strasbourg, France\\
        ICube, University of Strasbourg, CNRS, France\\
	\texttt{npadoy@unistra.fr} \\
}

\renewcommand{\headeright}{}
\renewcommand{\undertitle}{}
\renewcommand{\shorttitle}{MCIT-IG}

\hypersetup{
pdftitle={A template for the arxiv style},
pdfsubject={q-bio.NC, q-bio.QM},
pdfauthor={David S.~Hippocampus, Elias D.~Striatum},
pdfkeywords={First keyword, Second keyword, More},
}

\date{}

\begin{document}
\maketitle

\begin{abstract}
Surgical action triplets describe instrument-tissue interactions as \textlangle{instrument, verb, target}\textrangle{} combinations, thereby supporting a detailed analysis of surgical scene activities and workflow.
This work focuses on surgical action triplet {\it detection}, which is challenging but more precise than the traditional triplet {\it recognition} task as it consists of joint (1) localization of surgical instruments and (2) recognition of the surgical action triplet associated with every localized instrument.
Triplet detection is highly complex due to the lack of spatial triplet annotation.
We analyze how the amount of instrument spatial annotations affects triplet detection and observe that accurate instrument localization does not guarantee a better triplet detection due to the risk of erroneous associations with the verbs and targets.
To solve the two tasks, we propose \textbf{MCIT-IG}, a two-stage network, that stands for \textit{\textbf{M}ulti-\textbf{C}lass \textbf{I}nstrument-aware \textbf{T}ransformer - \textbf{I}nteraction \textbf{G}raph}. 
The \textbf{MCIT} stage of our network models per class embedding of the targets as additional features to reduce the risk of misassociating triplets. 
Furthermore, the \textbf{IG} stage constructs a bipartite dynamic graph to model the interaction between the instruments and targets, cast as the verbs. 
We utilize a mixed-supervised learning strategy that combines weak target presence labels for \textbf{MCIT} and pseudo triplet labels for \textbf{IG} to train our network. 
We observed that complementing minimal instrument spatial annotations with target embeddings results in better triplet detection.
We evaluate our model on the CholecT50 dataset and show improved performance on both instrument localization and triplet detection, topping the leaderboard of the CholecTriplet challenge in MICCAI 2022.
\end{abstract}


\section{Introduction}
Surgical workflow analysis in endoscopic procedures aims to process large streams of data~\cite{surgicalds} from the operating room (OR) to build context awareness systems~\cite{cai4cai}. These systems aim to provide assistance to the surgeon in decision making~\cite{deepcvs} and planning~\cite{lalys2014surgical}. Most of these systems focus on coarse-grained recognition tasks such as phase recognition~\cite{padoy2012}, instrument spatial localization and skill assessment~\cite{jin2018tool}.
Surgical action triplets~\cite{tripnet}, defined as \textlangle{}{\textit{instrument, verb, target}}\textrangle{}, introduce the fine-grained modeling of elements present in an endoscopic scene.
In cataract surgery,~\cite{lin2022instrument} adopts similar triplet formulation and also provides bounding box details for both instrument and targets. Another related work~\cite{bawa2021saras}, in prostatectomy, uses bounding box annotations for surgical activities defined as \textlangle{\textit{verb, anatomy}}\textrangle{}.
On laparoscopic cholecystectomy surgical data, existing approaches~\cite{rdv,sharma2022rendezvous,nwoye2022cholectriplet2021} focus on the challenging triplet recognition task, where the objective is to predict the presence of triplets but ignores their spatial locations in a video frame.

A recently conducted endoscopic vision challenge~\cite{nwoye2023cholectriplet2022} introduced the triplet detection task that requires instrument localization and its association with the triplet. 
While most of the contributed methods employ weak supervision to learn the instrument locations by exploiting the model's class activation map (CAM), a few other methods exploit external surgical datasets offering complementary instrument spatial annotations.
The significant number of triplet classes left the weakly supervised approaches at subpar performance compared to fully supervised methods. The results of the CholecTriplet2022 challenge~\cite{nwoye2023cholectriplet2022} have led to two major observations.

First, imprecise localization from the CAM-based methods impairs the final triplet detection performance, where the best weakly-supervised method reaches only $1.47\%$ detection mean average precision. 
Second, the correct association of triplet predictions and their spatial location is difficult to achieve with instrument position information alone. 
This is mainly due to the possible occurence of multiple instances of the same instrument and many options of targets/verbs that can be associated with one instrument instance.
We set two research questions following these observations:
(1) since manual annotation of instrument spatial locations is expensive and tedious, how can we use learned target/verb features to supplement a minimal amount of instrument spatial annotations?
(2) since instrument cues are insufficient for better triplet association, how can we generate valid representative features of the targets/verbs that do not require additional spatial labels?

To tackle these research questions, we propose a fully differentiable two-stage pipeline, \textbf{MCIT-IG}, that stands for \textit{\textbf{M}ulti-\textbf{C}lass \textbf{I}nstrument-aware \textbf{T}ransformer - \textbf{I}nteraction \textbf{G}raph}.
The MCIT-IG relies on instrument spatial information that we generate with Deformable DETR~\cite{zhu2021dd}, trained on a subset of Cholec80~\cite{endonet} annotated with instrument bounding boxes. In the first stage, MCIT, a lightweight transformer, learns class wise embeddings of the target influenced by instrument spatial semantics and high level image features. This allows the embeddings to capture global instrument association features useful in IG. We train MCIT with target binary presence label, providing weak supervision.
In the second stage, IG creates an interaction graph that performs dynamic association between the detected instrument instances and the target embeddings, and learns the verb on the interacting edge features, thereby detecting triplets. To train IG, triplet labels for the detected instrument instances are needed, which is unavailable. To circumvent this situation, we generate pseudo triplet labels for the detected instrument instances using the available binary triplet presence labels. In this manner, we provide mixed supervision to train MCIT-IG.

We hypothesize that a precise instrument detector can reveal additional instrument-target associations as more instrument instances are detected. 
To test this hypothesis, we conduct a study to investigate how the accuracy of the instrument detector affects triplet detection. We train an instrument detector with limited spatial data and evaluate the impact on triplet detection. We find that enhancing instrument localization is strongly linked to improved triplet detection performance. Finally, we evaluate our model on the challenge split of CholecT50~\cite{rdv,nwoye2023cholectriplet2022}. We report both improved instrument localization and triplet detection performance, thanks to the graph-based dynamic association of instrument instances with targets/verbs, that captures the triplet label.

\section{Methodology}

\begin{figure}[t!]
    \centering
    \includegraphics[width=1\linewidth]{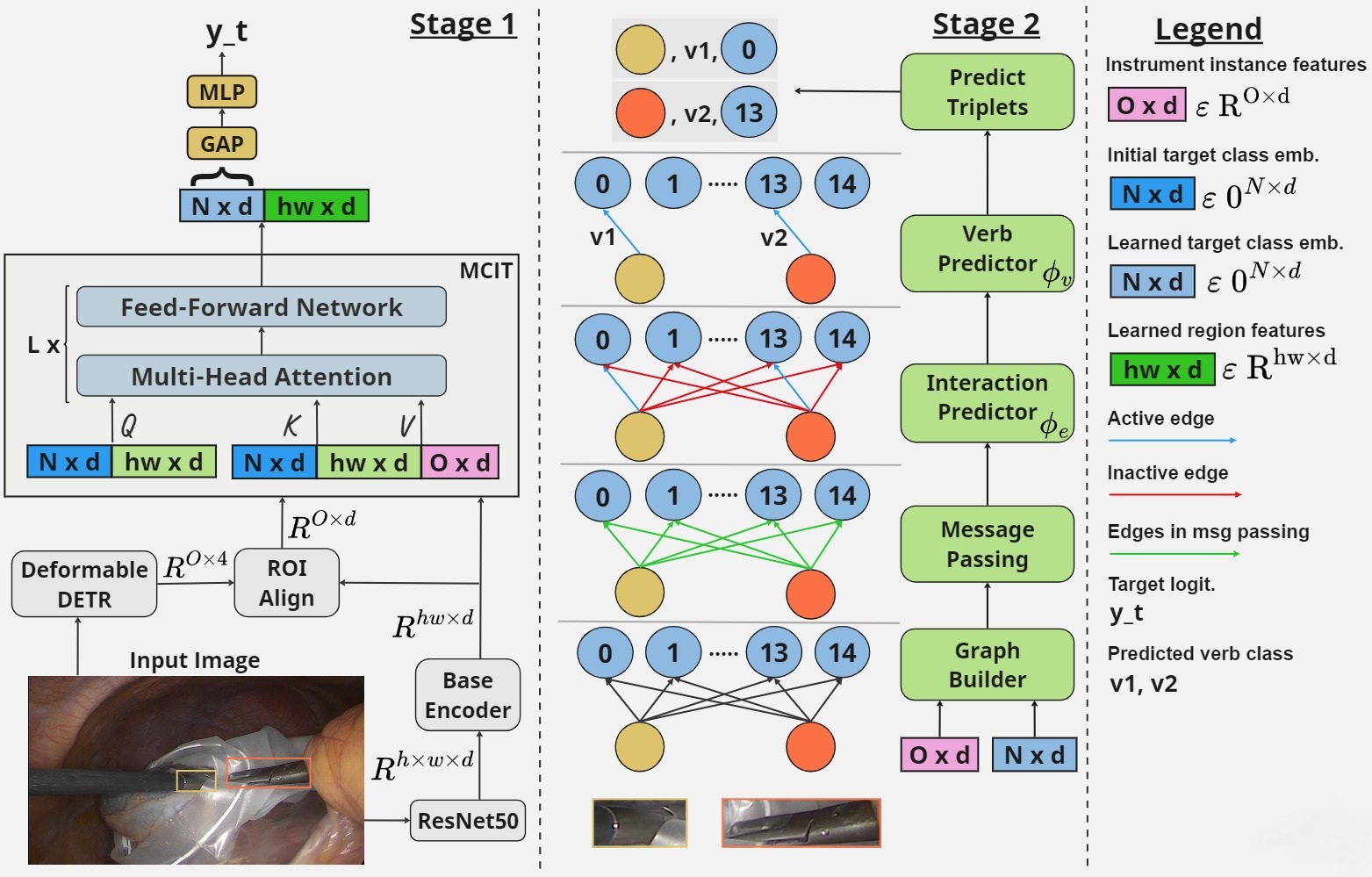}
    \caption{\it\textbf{Model Overview:} In Stage 1, \textbf{MCIT} learns instrument-aware target class embeddings.  In Stage 2, \textbf{IG} enforces association between instrument instances and target class embeddings and learns verb on the pairwise edge features. Green edges denotes all interactions and blue edges denotes active instrument-target pair with verb \textbf{v}. Red edges denotes no interaction.}
    \label{proposed_method}
\end{figure}

We design a novel deep-learning model that performs triplet detection in two-stages. In the first stage, we use a transformer to learn the instrument-aware target class embeddings. In the second stage, we construct an interaction graph from instrument instances to the embeddings, learn verb on the interacting edges and finally associate a triplet label with instrument instances. Using a trained Deformable DETR~\cite{zhu2021dd} based on the MMDetection~\cite{mmdetection} framework, we obtain bounding boxes for the instruments.


\noindent\paragraph{\textbf{Backbone:}}
To extract visual features, we utilize ResNet50~\cite{resnetcvpr16} as the backbone, and apply a $1\times1$ convolution layer to reduce the feature dimension from $\mathbb{R}^{h \times w \times c}$ to $\mathbb{R}^{h \times w \times d}$, where $c$ and $d$ are $2048$ and $512$ respectively. We flatten the features to $\mathbb{R}^{hw \times d}$ and input to the \textit{Base Encoder}, a lightweight transformer with $b_l$ layers. The base encoder modulates the local scene features from ResNet50 and incorporates context from other regions to generate global features $\mathcal{F}_{b} \in \mathbb{R}^{hw \times d}$ . We then apply \textit{ROIAlign} on $\mathcal{F}_{b}$ to generate instrument instance features $\mathcal{F}_{r} \in \mathbb{R}^{O \times d}$, where $O$ denotes the number of detected instruments. We also apply a linear layer $\Phi_{b}$ on the instrument box coordinates and concatenate with the embeddings of predicted instrument class category to get $d$-dimensional features, and finally fuse with $\mathcal{F}_{r}$ using linear layer $\Phi_{f}$ to produce final $d$-dimensional instrument features $\mathcal{F}_{i}$. 


\noindent\paragraph{\textbf{Learning Instrument-aware Target Class Embeddings:}}
To learn target features, we introduce a \textit{\textbf{M}ulti-\textbf{C}lass \textbf{I}nstrument-aware \textbf{T}ransformer (\textbf{MCIT})} that generates embeddings for each target class. In the standard transformer~\cite{vit}, a single class-agnostic token models the class distribution, but dilutes crucial class specific details. Inspired from~\cite{xu2022multi}, MCIT utilizes $N$ class tokens, $\mathcal{N}_{t} \in \mathbb{R}^{N \times d}$, to learn class-specific embeddings of the target. However, to make the class embeddings aware of the instruments in the scene, we use the instrument features $\mathcal{F}_{i}$ along with class tokens and region features.
Specifically, MCIT takes input ($\mathcal{F}_{b}$, $\mathcal{F}_{i}$) and creates learnable queries of dimension $\mathbb{R}^{(hw + N) \times d}$ and keys, values of dimension $\mathbb{R}^{(hw + N + O) \times d}$ to compute attention. Then, MCIT applies $t_l$ layers of attention to generate the output sequence, $\mathcal{P}_{t} \in \mathbb{R}^{(hw + N) \times d}$. The learned class embeddings are averaged across $N$ and input to a linear layer $\Phi_{t}$ to generate logits $y_{t} \in \mathbb{R}^{N}$ following Equation~\ref{target_logit}:
\begin{equation}
\label{target_logit}
y_{t} = \Phi_t \left( \frac{1}{N} \sum_{k=hw}^{hw+N} \mathcal{P}[k, :] \right).
\end{equation}
MCIT learns meaningful class embeddings of the target enriched with visual and position semantics of the instruments. This instrument-awareness is useful to identify the interacting instrument-target pairs. 

\noindent\paragraph{\textbf{Learning Instrument-Target Interactions:}}
To learn the interaction of the instrument and the target, we introduce a graph based framework \textit{\textbf{I}nteraction-\textbf{G}raph (\textbf{IG})} that relies on the discriminative features of instrument instances and target class embeddings. 
We create an unidirectional complete bipartite graph, $\mathcal{G} = (\mathcal{U},\mathcal{V},\mathcal{E})$, where $|\mathcal{U}| = O$ and $|\mathcal{V}| = N$ denotes the source and destination nodes respectively,
and edges $\mathcal{E} = \{e^{u}_{v}, {u} \in \mathcal{U} \wedge {v} \in \mathcal{V}\}$.
The node features of $\mathcal{U}$ and $\mathcal{V}$ correspond to the detected instrument instance features $\mathcal{F}_{i}$ and target class embeddings $\mathcal{N}_{t}$ respectively. We further project the nodes features to a lower dimensional space $d'$ using a linear layer $\Phi_{p}$. This setup provides an intuitive way to model instrument-tissue interactions as a set of active edges. Next, we apply message passing using GAT~\cite{velivckovic2017graph}, that aggregates instrument features in $\mathcal{U}$ and updates target class embeddings at $\mathcal{V}$.

\noindent\paragraph{\textbf{Learning Verbs:}}
We concatenate the source and destination node features of all the edges $\mathcal{E}$ in $\mathcal{G}$ to construct the edge feature $\mathcal{E}_{f} = \{e_{f}, e_{f} \in \mathbb{R}^{2{d}'}\}$. Then, we compute the edge confidence score $\mathcal{E}_{s} = \{e_{s}, e_{s} \in \mathbb{R}\}$ for all edges $\mathcal{E}$ in $\mathcal{G}$ by applying a linear layer $\Phi_{e}$ on $\mathcal{E}_{f}$. 
As a result, shown in Figure~\ref{proposed_method} Stage 2, only active edges (in blue) remain and inactive edges (in red) are dropped based on a threshold.
The active edge indicates the presence of interaction between an instrument instance and the target class. 
To identify the verb, we apply a linear layer $\Phi_{v}$ on $\mathcal{E}_{f}$ and generate verb logits $y_{v} \in \mathbb{R}^{V+1}$, where $V$ is the number of verb classes with an additional $1$ to denote background class. 

\noindent\paragraph{\textbf{Triplet Detection:}}
To perform target and verb association for each instrument instance $i$, first we select the active edge $e^{i}_{j}$ that corresponds to the target class $j = argmax(\alpha(\mathcal{E}^{i}_{s}))$, where $\alpha$ denotes softmax function and $\mathcal{E}^{i}_{s} = \{e^{u}_{s}, \forall e^{u}_{s} \in \mathcal{E}_{s} \wedge u = i \}$. For the selected edge ${e}' = e^{i}_{j}$, we apply softmax on the verb logits to obtain the verb class id, $k = argmax(\alpha(y^{{e}'}_{v}))$. The final score for the triplet \textlangle{}{\textit{i, k, j}}\textrangle{} is given by $p({e^{i}_{j}}) \times p({y^{{e}'}_{v_{k}}})$, where $p$ denotes the probability score. 

\noindent\paragraph{\textbf{Mixed Supervision:}}
We train our model in two stages. In the first stage, we train MCIT to learn target classwise embeddings with target binary presence label with weighted binary cross entropy on target logits $y_t$ for multi-label classification task following Equation~\ref{bce_formula}:
\begin{equation}
\label{bce_formula}
L_t = \sum_{c=1}^{C}\frac{-1}{N}\left( W_cy_clog\left(\sigma(\hat{y}_c)\right) + (1-y_c)log\left((1-\sigma(\hat{y}_c)\right) \right),
\end{equation}
where $C$ refers to total number of target classes, $y_{c}$ and $\hat{y}_{c}$ denotes correct and predicted labels respectively, $\sigma$ is the sigmoid function and $W_{c}$ is the class balancing weight from~\cite{rdv}.
For the second stage IG, we generate pseudo triplet labels for each detected instrument instances, where we assign the triplet from the binary triplet presence label if the corresponding instrument class matches.
To train IG, we apply categorical cross entropy loss on edge set $\mathcal{E}^{i}_{s}$ and verb logits $y^{{e}'}_{v}$ for all instrument instances $i$ to obtain losses $L_\mathcal{G}^e$ and $L_\mathcal{G}^v$ respectively following Equation~\ref{ce_formula}:
\begin{equation}
\label{ce_formula}
L = -\sum_{c=1}^My_{c}\log(p_{c}),
\end{equation}
where $M$ denotes the number of classes which is $N$ for $L_\mathcal{G}^e$ and $V+1$ for $L_\mathcal{G}^v$. The final loss for training follows Equation~\ref{final_loss}:
\begin{equation}
\label{final_loss}
L = L_t + \alpha \times L_\mathcal{G}^e + \beta \times L_\mathcal{G}^v,
\end{equation}
where $\alpha$ and $\beta$ denote the weights to balance the loss contribution.

\section{Experimental Results and Discussion}

\subsection{Dataset \& Evaluation Metrics}
Our experiments are conducted on the publicly available CholecT50~\cite{rdv} dataset, which includes binary presence labels for $6$ instruments, $10$ verbs, $15$ targets, and $100$ triplet classes. We train and validate our models on the official challenge split of the dataset~\cite{ctsplits}. The test set consists of $5$ videos annotated with instrument bounding boxes and matching triplet labels.
Since the test set is kept private to date, all our results are obtained by submitting our models to the challenge server for evaluation.
The model performance is accessed using video-specific average precision and recall metrics at a threshold ($\theta=0.5$) using the ivtmetrics library~\cite{ctsplits}.
We also provide box association results in the supplementary material for comparison with other methods on the challenge leaderboard.

\subsection{Implementation Details}
We first train our instrument detector for $50$ epochs using a spatially annotated $12$ video subset of Cholec80 and generate instrument bounding boxes and pseudo triplet instance labels for CholecT50 training videos.
In stage 1, we set $b_l = 2$, $t_l = 4$, and $d$ to $512$. We initialize target class embeddings with zero values. We use 2-layer MLP for $\Phi_{b}$, $\Phi_{f}$, and 1-layer MLP for $\Phi_{t}$. We resize the input frame to $256 \times 448$ resolution and apply flipping as data augmentation. For training, we set learning rate $1e^{-3}$ for (backbone, base encoder), and $1e^{-2}$ for MCIT. We use SGD optimizer with weight decay $1e^{-6}$ and train for $30$ epochs.
To learn the IG, we fine-tune stage 1 and train stage 2. We use learning rate $1e^{-4}$ for (MCIT, base encoder), and $1e^{-5}$ for the backbone. In IG , $\Phi_{p}$, $\Phi_{e}$, and $\Phi_{v}$ are 1-layer MLP with learning rate set to $1e^{-3}$, and $d'$ set to $128$ in $\Phi_{p}$ to project node features to lower dimensional space. We use Adam optimizer and train both stage 1 and stage 2 for $30$ epochs, exponentially decaying the learning rate by $0.99$. The loss weights $\alpha$ and $\beta$ is set to 1 and 0.5 respectively. We set batch size to $32$ for both stages. We implement our model in PyTorch and IG graph layers in DGL~\cite{wang2019dgl} library. We train the model on Nvidia V100 and A40 GPUs and tune model hyperparameters using random search on $5$ validation videos.

\subsection{Results}
\noindent\paragraph{\textbf{Comparison with the baseline:}}
We obtain the code and weights of Rendezvous (RDV)~\cite{rdv} model from the public github and generate the triplet predictions on the CholecT50-challenge test set. We then associate these predictions with the bounding box predictions from the Deformable DETR~\cite{zhu2021dd} to generate baseline triplet detections as shown in Table \ref{mainresult}. 
With a stable instrument localization performance ($60.1$ mAP), our MCIT model leverage the instrument-aware target features to captures better semantics of instrument-target interactions than the baseline. Adding IG further enforces the correct associations, thus improving the triplet detection performance by $+0.89$ mAP, which is $13.8\%$ increase from the baseline performance. Also, the inference time in frame per seconds (FPS) for our MCIT-IG model on Nvidia V100 is $25$ compared to $28.1$ in RDV.

\begin{table}[tb]
\centering
    \setlength{\tabcolsep}{4pt}
    \caption{\label{mainresult}Results on Instrument Localization and Triplet Detection (mAP@0.5 in \%).}
    \resizebox{\textwidth}{!}{%
        \begin{tabular}{@{}lcccccccccr@{}}\toprule
            \multirow{2}{*}{Method}&\phantom{abc}&
            \multirow{2}{*}{Instrument Detector}&\phantom{abc}&
            \multicolumn{3}{c}{Instrument localization}&\phantom{abc}&
            \multicolumn{3}{c}{Triplet detection}\\ \cmidrule{5-7} \cmidrule{9-11} 
             &&&& $AP_{I}$ & & $AR_{I}$ && $AP_{IVT}$ &  & $AR_{IVT}$ \\ \midrule
             RDV (Baseline)~\cite{rdv} &&Deformable DETR \cite{zhu2021dd}&& \textbf{60.1} & & \textbf{66.6} && 6.43 &  & 9.50 \\
             MCIT (Ours)&&Deformable DETR~\cite{zhu2021dd}&& \textbf{60.1} & & \textbf{66.6} && 6.94 &  & 9.80 \\
             MCIT+IG (Ours)&&Deformable DETR~\cite{zhu2021dd}&& \textbf{60.1} & & \textbf{66.6} && \textbf{7.32} &  & \textbf{10.26} \\
            \bottomrule            
        \end{tabular}
        }
\end{table}

\noindent \paragraph{\textbf{Ablation Study on the Spatial Annotation Need:}}
Here, we study the impact of an instrument localization quality on triplet detection and how the target features can supplement fewer spatial annotations of the instruments for better triplet detection.
We compare with ResNet-CAM-YOLOv5~\cite{nwoye2023cholectriplet2022} and Distilled-Swin-YOLO~\cite{nwoye2023cholectriplet2022} models which were also trained with bounding box labels.
We observed that the triplet detection mAP increases with increasing instrument localization mAP for all the models as shown in Table \ref{detector_perf}. 
However, the scale study shows that with lesser bounding box instances, our MCIT-IG model stands tall:
outperforming Distilled-Swin-YOLO by +0.86 mAP with ${\sim}9K$ fewer frames and surpassing ResNet-CAM-YOLOv5 by +1.42 mAP with ${\sim}7K$ frames to spare.
Note that a frame can be annotated with one or more bounding boxes.

\begin{table}[tp]
\centering
    \setlength{\tabcolsep}{5pt}
    \caption{\label{detector_perf}Instrument Localization and Triplet Detection (mAP@0.5 in $\%$) vs number of videos$\slash$frames used in training the Instrument Detector.}
    \resizebox{\textwidth}{!}{%
        \begin{tabular}{@{}lccccccccr@{}}\toprule
            \multirow{2}{*}{Method}&\multirow{2}{*}{\#Videos}&\multirow{2}{*}{\#Frames}\phantom{abc}&
            \multicolumn{3}{c}{Instrument localization}&\phantom{abc}&
            \multicolumn{3}{c}{Triplet detection}\\ \cmidrule{4-6} \cmidrule{8-10}
             &&& $AP_{I}$ & & $AR_{I}$ && $AP_{IVT}$ &  & $AR_{IVT}$ \\ \midrule
             ResNet-CAM-YOLOv5 \cite{nwoye2023cholectriplet2022} &51&${\sim}$22000 & 41.9 & & 49.3 && 4.49 &  & 7.87 \\
             Distilled-Swin-YOLO \cite{nwoye2023cholectriplet2022} &33&${\sim}$13000 & 17.3 & & 30.4 && 2.74 &  & 6.16 \\ \midrule
             MCIT+IG (Ours) &1  &4214  & 33.5 & & 39.6 && 3.60 &  & 4.95 \\
             MCIT+IG (Ours) &5  &15315 & 53.1 & & 59.6 && 5.91 &  & 8.73 \\
             MCIT+IG (Ours) &12 &24536 & \textbf{60.1} & & \textbf{66.6} && \textbf{7.32} &  & \textbf{10.26} \\
            \bottomrule
        \end{tabular}
        }
\end{table}
\noindent \paragraph{\textbf{Ablation Studies on the Components of MCIT-IG:}}
We analyze the modules used in MCIT-IG and report our results in Table~\ref{compresult}. Using both ROI and box features provides a complete representation of the instruments that benefits IG, whereas using just ROI or box features misses out on details about instruments hurting the triplet detection performance.
We further test the quality of target class embeddings without instrument awareness in MCIT. Results in Table~\ref{compresult} indicates that the lack of instrument context hampers the ability of the target class embeddings to capture full range of associations with the triplets.
Also, message passing is key in the IG as it allows instrument semantics to propagate to target class embeddings, which helps distinguish interacting pairs from other non-interacting pairs.

\begin{table}[tp]
\centering
    \setlength{\tabcolsep}{20pt}
    \caption{\label{compresult}Component-wise performance on Triplet Detection (mAP@0.5 in \%).}
    \resizebox{\textwidth}{!}{%
        \begin{tabular}{@{}lccccccr@{}}\toprule
             & $ROI$ & $Box$ & $Graph$ & $Tool Awareness$ & $AP_{IVT}$ & $AR_{IVT}$ \\ \midrule
             & \checkmark &  & \checkmark &\checkmark & 4.11 & 5.64 \\
             &  & \checkmark & \checkmark &\checkmark & 4.97 & 6.93 \\
             & \checkmark & \checkmark & \checkmark &  & 4.98 & 7.71 \\
             & \checkmark & \checkmark &  &\checkmark & 6.94 & 9.80 \\
             & \checkmark & \checkmark & \checkmark &\checkmark & \textbf{7.32} & \textbf{10.26} \\
            \bottomrule
        \end{tabular}
        }
\end{table}

\begin{table}[tp]
    \centering
    \setlength{\tabcolsep}{3pt}
    \caption{\label{sotaresult}Comparison with top methods from CholecTriplet 2022 Challenge~\cite{nwoye2023cholectriplet2022}, leaderboard results on \color{blue}\url{ https://cholectriplet2022.grand-challenge.org/results}.}
    \resizebox{\textwidth}{!}{%
    \begin{tabular}{@{}lcccrccrcc@{}}
        \toprule
        \multirow{2}{*}{Method} &
        \multirow{2}{*}{Params (M)} &
        \multirow{2}{*}{Supervision} &
        \multirow{2}{*}{Ranking} &
        \phantom{abc} & \multicolumn{2}{c}{Instrument localization} &
        \phantom{abc} & \multicolumn{2}{c}{Triplet detection}\\
         \cmidrule{6-7}\cmidrule{9-10} 
             &&&&& $AP_{I}$ & $AR_{I}$ && $AP_{IVT}$ & $AR_{IVT}$ \\ \midrule
             RDV-Det & 17.1 & Weak & 5$^{th}$ && 3.0 & 7.6 && 0.24 & 0.79 \\
             DualMFFNet & 28.3 & Weak & 4$^{th}$ &&  4.6 & 6.6 && 0.36 & 0.73 \\
             MTTT & 181.7 & Weak & 3$^{rd}$ &&  11.0 & 21.1 && 1.47 & 3.65 \\
             Distilled-Swin-YOLO & 88 & Full & 2$^{nd}$ && 17.3 & 30.4 && 2.74 & 6.16 \\
             ResNet-CAM-YOLOv5 & 164 & Full & 1$^{st}$ && 41.9 & 49.3 && 4.49 & 7.87 \\
             MCIT+IG (Ours) & 100.6& Mixed & -- && \textbf{60.1} & \textbf{66.6} && \textbf{7.32} & \textbf{10.26} \\
            \bottomrule
    \end{tabular}
    }
\end{table}

\noindent\paragraph{\textbf{Comparison with the State-of-the-art (SOTA) Methods:}}
Results in Table~\ref{sotaresult} show that our proposed model outperforms all the existing methods in the CholecTriplet 2022 challenge \cite{nwoye2023cholectriplet2022}, obtained the highest score that would have placed our model $1^{st}$ on the challenge leaderboard in all the accessed metrics.
Leveraging our transformer modulated target embeddings and graph-based associations, our method shows superior performance in both instrument localization and triplet detection over methods weakly-supervised on binary presence labels and those fully supervised on external bounding box datasets like ours. 
More details on the the challenge methods are provided in~\cite{nwoye2023cholectriplet2022}.

\section{Conclusion}
In this work, we propose a fully differentiable two-stage pipeline for triplet detection in laparoscopic cholecystectomy procedures. We introduce a transformer-based method for learning per class embeddings of target anatomical structures in the absence of target instance labels, and an interaction graph that dynamically associates the instrument and target embeddings to detect triplets. We also incorporate a mixed supervision strategy to help train MCIT and IG modules. We show that improving instrument localization has a direct correlation with triplet detection performance. We evaluate our method on the challenge split of the CholecT50 dataset and demonstrate improved performance over the leaderboard. 

\paragraph{Acknowledgements.}
This work was supported by French state funds managed by the ANR within the National AI Chair program under Grant ANR-20-CHIA-0029-01 (Chair AI4ORSafety) and within the Investments for the future program under Grant ANR-10-IAHU-02 (IHU Strasbourg). It was also supported by BPI France under reference DOS0180017/00 (project 5G-OR). It was granted access to the HPC resources of Unistra Mesocentre and GENCI-IDRIS (Grant AD011013710).

\bibliographystyle{IEEEtran}
\bibliography{references}

\newpage
\renewcommand{\thesubsection}{\Alph{subsection}}

\section*{Additional Results}\label{extra_results}
\begin{table}[ht]
\centering
    \setlength{\tabcolsep}{10pt}
    \caption{\label{breakdown1}Model evaluation performance compared with CholecTriplet2022 challenge submissions~\cite{nwoye2023cholectriplet2022} on triplet association metrics (TAS)~\cite{ctsplits}. Our method improves on the \textit{LM} metric (\% of tools localized at given IoU and matched with correct triplet) over the leaderboard by $+5.7\%$.}
    \resizebox{\textwidth}{!}{%
        \begin{tabular}{@{}lcccccccr@{}}\toprule
             Method & $LM\uparrow$ & $pLM$ & $IDS\downarrow$ & $IDM\downarrow$ & $MIL\downarrow$ & $RFP\downarrow$ & $RFN\downarrow$ & $AP_{IVT}$ \\ \midrule
             MCIT+IG (Ours)      & \textbf{29.6}  &\textbf{3.4}    & 0.5 & 0.1 & \textbf{0.0} & 4.5 & 61.9 & \textbf{7.32}\\
             ResNet-CAM-YOLOv5   & 23.9   & 8.2   & 0.9            & 0.1   & 0.1  & \textbf{3.3}    & 63.5 & 4.49\\
             Distilled-Swin-YOLO & 12.0   & 11.3  & 0.3            & \textbf{0.0}   & 0.3  & 33.0   & 43.1 & 2.74\\
             MTTT                & 8.6    & 25.4  & 0.1            & 0.1   & 0.4  & 11.6   & 53.8 & 1.47\\
             RDV-Det             & 3.3    & 29.0  & \textbf{0.0}   & \textbf{0.0}   & 1.1  & 5.4    & 61.1 & 0.24 \\
             DualMFFNet          & 3.0    & 17.4  & 0.1            & \textbf{0.0}   & 1.2  & 3.5    & 74.8 & 0.36\\
             IF-Net              & 2.0    & 14.9  & 0.1            & 0.7    & 1.5  & 35.9   & 45.1 & 0.22\\
             SurgNet             & 2.0    & 8.5   & \textbf{0.0}   & \textbf{0.0}   & 1.6  & 18.9   & 69.0 & 0.13\\
             DATUM               & 0.4    & 22.8  & \textbf{0.0}   & \textbf{0.0}   & 1.3  & 45.1   & \textbf{30.5} & 0.08\\
             AtomTKD             & 0.1    & 14.9  & \textbf{0.0}   & \textbf{0.0}   & 3.1  & 28.4   & 53.6 & 0.15\\
            \bottomrule
        \end{tabular}
        }
\end{table}

\begin{figure}[ht]
    \centering
    \includegraphics[width=1.0\linewidth]{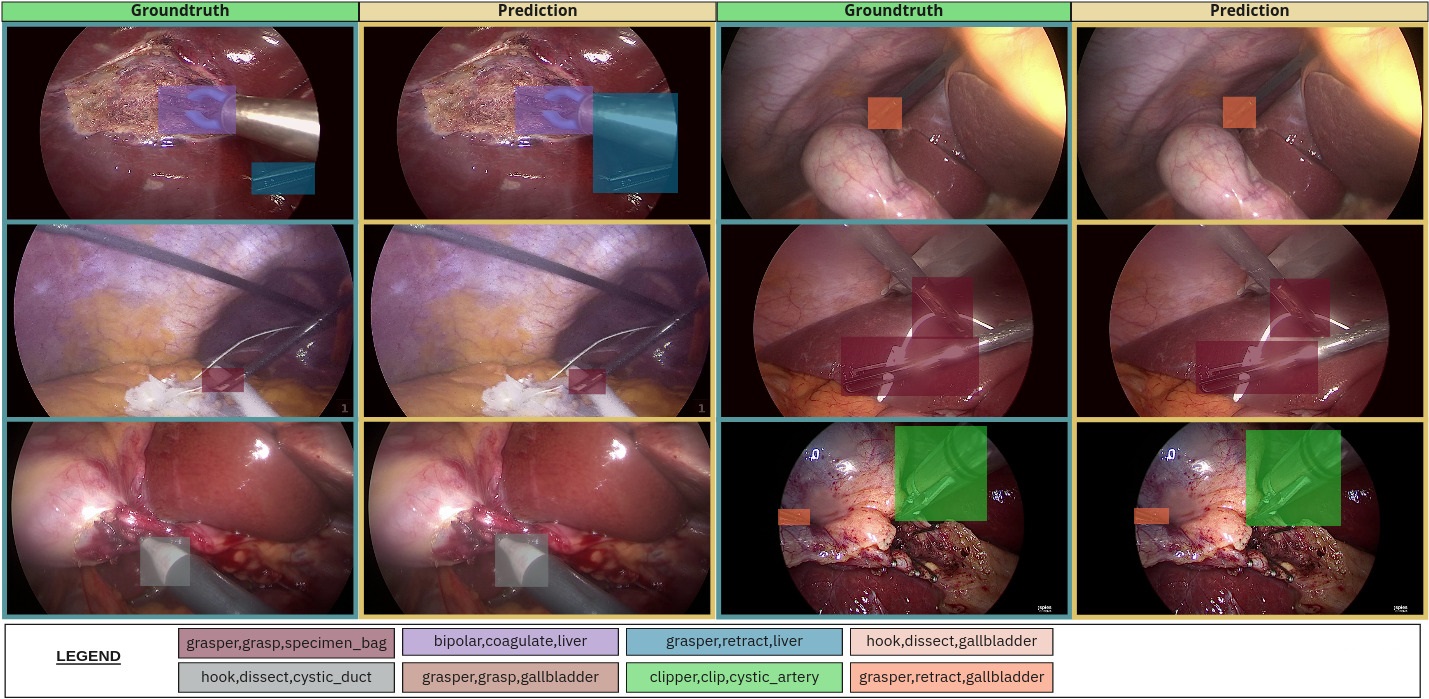}
    \caption{\it\textbf{Qualitative Results:} MCIT-IG predictions on random images from the validation set. Our trained Instrument Detector is able to correctly localize the instruments and improve on the triplet detection. Please zoom for more details.(Best viewed in color).}
    \label{proposed_method}
\end{figure}

\begin{table}[ht]
\centering
    \setlength{\tabcolsep}{12pt}
    \caption{\label{ig_result}Comparison of different message passing methods in Interaction Graph (IG) for Triplet Detection (mAP@0.5 in \%).}
    \resizebox{\textwidth}{!}{%
        \begin{tabular}{@{}lcccccr@{}}\toprule
             & $Message Passing$ & $NumHeads$ & $Feature Dimension$ & $NumLayers$ & $AP_{IVT}$ & $AR_{IVT}$ \\ \midrule
             & GCN~\cite{kipf2016semi} & - & 128 & 2 & 5.54 & 8.25\\
             & SAGE~\cite{hamilton2017inductive} & - & 128 & 2 & 5.67 & 8.05\\
             & GAT~\cite{velivckovic2017graph} & 2 & 128 & 2 & \textbf{7.32} & \textbf{10.26}\\
            \bottomrule
        \end{tabular}
        }
\end{table}

\newpage
\begin{table}[ht]
\centering
    \setlength{\tabcolsep}{42pt}
    \caption{\label{breakdown1}Training hyper-parameters of Deformable DETR~\cite{zhu2021dd} for the Instrument Detector.}
    \resizebox{\textwidth}{!}{%
        \begin{tabular}{@{}lc||cc@{}}\toprule
             Backbone      & ResNet50 & LR Scheduler & step  \\
             Warmup        & linear   & Batch Size    & 32    \\
             Warmup Iterations  & 500 & Optimizer     & AdamW \\
             Warmup Ratio  & 0.001    & Weight Decay  & 1e-4  \\
             NumEpochs     & 50       & nGPUs         & 4     \\
             Learning Rate & 2e-4    \\
            \bottomrule
        \end{tabular}
        }
\end{table}


\end{document}